\pdfoutput=1

\documentclass[11pt]{article}

\usepackage[final]{acl}

\usepackage{times}
\usepackage{latexsym}

\usepackage[T1]{fontenc}

\usepackage[utf8]{inputenc}

\usepackage{microtype}

\usepackage{inconsolata}

\usepackage{graphicx}

\usepackage{float}
\usepackage{multirow}
\usepackage{amsmath}
\usepackage{soul}
\usepackage{makecell}
\usepackage{pifont}
\newcommand{\cmark}{\ding{51}}%
\newcommand{\xmark}{\ding{55}}%
\usepackage{float}
\usepackage{gb4e}\noautomath
\usepackage{makecell}
\usepackage{linguex} 
\usepackage[tone]{tipa}
\let \ipa \textipa
\usepackage{booktabs}

\usepackage[nameinlink,capitalize]{cleveref}

\newif\ifshownotes
\shownotesfalse 
\newcommand{\isabel}[1]{\ifshownotes{\small \textcolor{orange}{[Isabel: #1]}}\fi}

\title{Investigating the interaction of linguistic and mathematical reasoning in language models using multilingual number puzzles}

\author{
  \textbf{Antara Raaghavi Bhattacharya\textsuperscript{1, 2}},
  \textbf{Isabel Papadimitriou\textsuperscript{3}}, \\
  \textbf{Kathryn Davidson\textsuperscript{2}},
  \textbf{David Alvarez-Melis\textsuperscript{1,3}}
\\
\\
  \textsuperscript{1}School of Engineering and Applied Sciences, Harvard University\\ 
  \textsuperscript{2}Department of Linguistics, Harvard University\\
  \textsuperscript{3}Kempner Institute for the Study of Natural and Artificial Intelligence at Harvard University \\
  \small{
    \textbf{Correspondence:} \texttt{\href{mailto:antara@alumni.harvard.edu}{antara@alumni.harvard.edu}}
  }
}

\begin{document}
\maketitle
\begin{abstract}
    

Across languages, numeral systems vary widely in how they construct and combine numbers. While humans consistently learn to navigate this diversity, large language models (LLMs) struggle with linguistic-mathematical puzzles involving cross-linguistic numeral systems, which humans can learn to solve successfully. We investigate \textit{why} this task is difficult for LLMs through a series of experiments that untangle the linguistic and mathematical aspects of numbers in language. Our experiments establish that models cannot consistently solve such problems unless the mathematical operations in the problems are explicitly marked using known symbols ($+$, $\times$, etc, as in ``twenty + three''). In further ablation studies, we probe how individual parameters of numeral construction and combination affect performance. While humans use their linguistic understanding of numbers to make inferences about the implicit compositional structure of numerals, LLMs seem to lack this notion of implicit numeral structure. We conclude that the ability to flexibly infer compositional rules from implicit patterns in human-scale data remains an open challenge for current reasoning models.

\end{abstract}

\section{Introduction}
Language models reason and solve problems using language. What is the connection (and the integration) between their linguistic systems and their impressive reasoning abilities? To investigate this question, we run a suite of experiments to analyze how language models solve puzzles about diverse linguistic number systems. People represent numbers through language, using rule-based systems that are simultaneously linguistic and mathematical \cite{ifrah2000universal, dehaene2011number, carey2004bootstrapping, le2007one, ionin2006composition, hammarstrom2010rarities, comrie2011typology}. Unlike most mathematical reasoning problems, where the mathematical operators are explicit, a numeral system contains implicit operations for describing numerals, and there is considerable variety in how this is done across the world's languages. For example, French \textit{vingt-neuf} (20 + 9), Bengali \textit{untir\={\i}sh} (30 $-$ 1), Tamil \textit{\ipa{irupatti o\=*npatu}} (2 $\times$ 10 $+$ (10 - 1)), and Birom \textit{\ipa{bākūrū bībā ná vE tùNūn}} (2 $\times$ 12 $+$ 5) all evaluate to the Hindu-Arabic numeral 29.



\begin{table*}[t]
\centering
{\fontfamily{lmss}\selectfont
\begin{tabular}{lllll}
\toprule
\multicolumn{1}{l}{\textbf{Variable}} & \multicolumn{3}{l}{\textbf{Operator}} & \textbf{Example} \\
& Explicitness & Familiarity & Type & \\
\midrule
Single character & Implicit & - & - & A B \\
 & Explicit & Familiar & Symbol &  A + B \\
 & Explicit & Unfamiliar & Symbol &  A $\alpha$ B \\
 & Explicit & Unfamiliar & Word &  A \textit{xebrut} B \\
Multi-character & Implicit & - & - & gbaifi pagig \\
& Explicit & Familiar & Symbol & gbaifi + pagig \\
\textbf{\dots} & \textbf{\dots} & \textbf{\dots} & \textbf{\dots}  \\
\bottomrule
\end{tabular}
}
\caption{A demonstration of the experimental conditions for our explicit operators experiment. We add explicit operators to our base \textsc{implicit} problems, using both the familiar symbols for addition/multiplication/subtraction, as well as unfamiliar symbols and words to symbolize the operation.}
\label{tab:operation-experiment}
\end{table*}

We investigate the capabilities of language models to solve puzzles about linguistic number systems, drawn from linguistics competitions (Linguistics Olympiads) where high-school students have to reason through data about unknown languages and explain the linguistic rules governing the data \cite{derzhanski2010linguistics}. While language models approach human performance on several language-based benchmarks \cite{hendrycks2020measuring, kojima2022large, beguvs2023large}, and recent reasoning models deliberately optimized for logical and mathematical reasoning show remarkable performance improvements for many structured mathematical reasoning tasks \cite{zhong2024evaluation, jaech2024openai}, LLMs perform extremely poorly at solving linguistic-mathematical puzzles about systems of numbers in different languages \cite{derzhanski2018linguistic, bean2024lingoly}. 



\textbf{Why do language models fail to solve these problems at the intersection of language and math} --- what specifically causes this failure? And how much of this failure is due to the linguistic vs. the mathematical aspects of the problem? 

We present a method to systematically isolate individual parameters of number construction and combination and investigate how they affect language model performance. We establish that most individual mathematical features (like base) do not hinder the ability of sufficiently advanced language models to solve such problems. However, unless \textbf{the mathematical operations in a problem are made explicit through familiar symbols (+, ×, etc.)}, models cannot consistently solve the problem. This indicates that, at least within the domain of linguistic-mathematical problems, models cannot infer the compositional structure of numerals like humans can, or sufficiently abstract notions like operators. We discuss our findings in the broader context of human language, concluding that flexible, adaptive use of language across domains appears to remain challenging for LLMs.



\begin{figure*}[ht]

  \centering
  \includegraphics[width=0.8\linewidth]{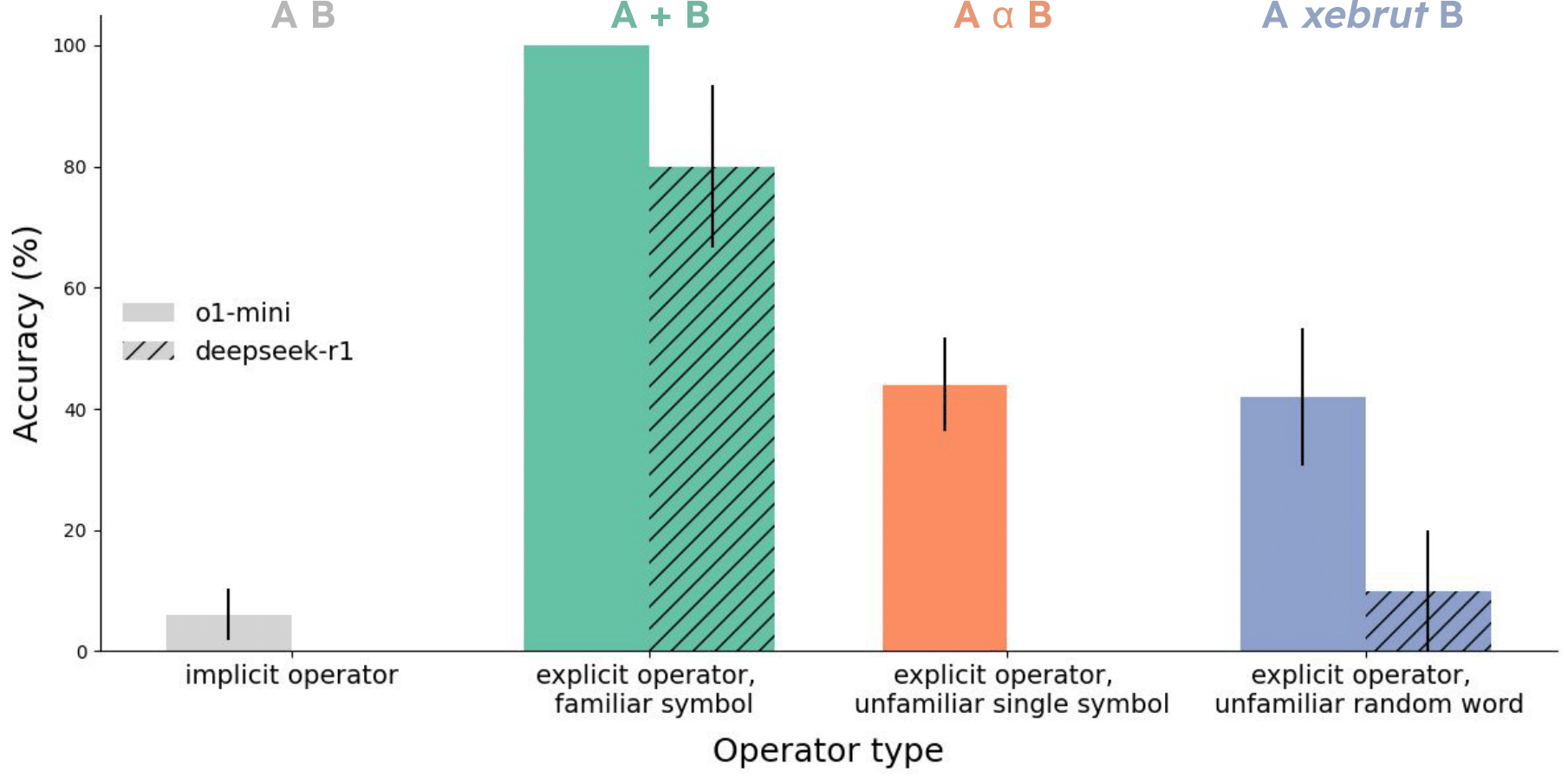}
  
  \caption {\textbf{Making operators explicit significantly improves performance}. Results for explicit operator experiments, for the single-character variable case (For the results on multi-character variables, see \Cref{app:multi_char} \Cref{fig:multitok_delta}). Making operators explicit shows performance improvement over the \textsc{implicit} condition, but this is only substantially and reliably the case when the operator is made explicit with a familiar symbol like ``+''.  Error bars denote standard error of the mean. 10 problems, 5 iterations per problem.}

  \label{fig:1}
\end{figure*}

\section{Background}

\subsection{Linguistic and cognitive connections}


People acquire systems of number representation as part of learning language, and are consequently able to construct arbitrary numerals using the rules that they learn. Although the system of rules may be language-specific, the general framework of numeral construction and combination is a fundamental cognitive ability \cite{hurford1987language, feigenson2004core}. Performing mathematics in a symbolic sense requires explicit instruction (e.g. a child would not inherently know what + connotes), but once this symbolic meaning has been learned, people can generalize it to apply to any numbers \cite{sarnecka2015counting}.

Numeral operations in language can be marked both explicitly (e.g. \textit{und} in German \textit{einundzwanzig}) and implicitly (as in English \textit{twenty-one}), with larger numerals often using a combination of implicit and explicit operations (\textit{five hundred and one} = 5 [$\times$] 100 + 1). Even when operations are implicit, people can understand and infer the cross-linguistic compositional structure of numerals \cite{ionin2006composition}.

In a linguistics contest, a high-school student would not need to know any mathematical concepts beyond basic arithmetic to reason through number system problems and infer the rules needed to solve them. The challenge lies instead in whether models can learn and infer such rules from limited data — a characteristic capacity of humans acquiring language.

\subsection{Mathematical ability in language models}


Recent language models seem to display strong numerical understanding and processing abilities if presented with purely mathematical problems in standard formats \cite{yang2024number}, particularly for small numbers and simple mathematical operations (of the kinds used in linguistics contest problems). Current reasoning models appear to perform well at arithmetic and algebra, math word problems \cite{ahn2024large}, and difficult mathematical contest questions equivalent to advanced college-level math problems \cite{fang2024mathodyssey, chervonyi2025gold}, although their problem-solving ability is sometimes inconsistent \cite{mccoy2023embers, shojaee2025illusion}. If such models are unable to solve linguistic-mathematical problems involving much simpler mathematics, and introducing linguistic structure into the problem causes their reasoning ability to break down, this indicates limitations in the \textit{scope} of their reasoning — models may be unable to apply their reasoning flexibly across domains in the ways that humans do.

\section{Methods}

\textbf{Models.} We used OpenAI o1-mini \cite{jaech2024openai} and DeepSeek-R1-distill-Qwen-7B \cite{guo2025deepseek} reasoning models to conduct our experiments, querying o1-mini via the API and running DeepSeek locally. All code and data used for our experiments are available at 
\url{https://github.com/antara-raaghavi/multilingual-number-puzzles}. 

We additionally queried an instruction-tuned model (qwen-2-7b-vl-instruct) and a base model (llama-3.1-8B), both of which had an accuracy of 0 across all conditions that we test. These models almost always generated longer text answers without numbers rather than the simple numerical answer required, and were hence excluded from our analyses.

\textbf{Data.}  We obtained data for linguistics olympiad problems from two publicly available datasets: LingOly \cite{bean2024lingoly} and Linguini \cite{sanchez2024linguini}, filtering both datasets for problems tagged as “number systems”. After filtering, we had 15 problems from the LingOly and 8 problems from the Linguini dataset. Not every problem in the dataset could be standardized in the ways that our experiments required. The entire dataset was thus manually evaluated for suitable problems, and 10 problems were chosen for evaluation, all in distinct languages (see \Cref{app:lang-table}). These problems spanned a range of difficulty from the first round of the UK Linguistics Olympiad to the International Linguistics Olympiad (most challenging).

\section{Experiments}

\subsection{The effect of explicit operators in problems}
\label{expt:1}

Since so many of the mathematical operators in numeral structure are implicit (e.g. in English we say `twenty three' to mean `twenty + three'), our first experiment investigates how this implicit structure affects how models solve the problems. To do this, we standardize and convert the 10 existing linguistic number system problems to  
mathematical problems, and vary how explicit the operators are, as shown in \Cref{tab:operation-experiment}.

First, we standardize all problems to control for model tokenization and task-external knowledge effects: we identify all meaningful morphemes, standardize all phonological changes, and replace them with  dummy words as described in detail in \Cref{subsec:randomize}. This standardized version of each problem is what we call the \textsc{implicit} setting, since the mathematical operations are largely implicit, as they are in language. Taking these \textsc{implicit} problems as our baselines, we then make the operators explicit in three ways: 1) as the familiar mathematical operator symbols that perform the operation (e.g. `+' for addition), 2) as symbols that are unfamiliar for performing that operation, and 3) as whole words sampled from the tokenizer. A full example prompt with a puzzle in four variations is provided in \Cref{app:multi_char}.

We present our results in \Cref{fig:1}. In all cases, the presence of explicit operations with familiar symbols yields significant improvements over the default \textsc{implicit} condition (o1-mini performs at ceiling). In the multi-character setting (more linguistic), models perform better on average in the \textsc{implicit} condition than in the case with an explicit operator as an unfamiliar random word (vid. \Cref{fig:multitok_delta}). It is likely harder to differentiate between function words (operators) and number words (numerals) in such a setting --- this finding is consistent with work that has shown human solvers also find a problem to be more difficult when the operator word is explicit but unfamiliar \cite{derzhanski2018linguistic}. Overall, our results demonstrate that it is difficult for models to reason about the abstract idea that linguistic quantities might contain \textit{operators}, if the operators are not explicitly provided using familiar symbols. 


\subsubsection{Error analysis}



We observe some common patterns of error in the model responses. For the three problems which involved squares and cubes of numbers, when the operators were not explicit and familiar, o$1$-mini almost always responded by pattern-matching (e.g. providing another square/cube number) instead of solving the problem, as seen in \Cref{tab:error-analysis}. o$1$-mini also reproduced a number given in the input question as the answer in several cases (11 for the multi-token condition, and 3 for the single-token condition, across 150 trials) when the operators were not explicit and familiar.  

\begin{table}[h]
\centering
\begin{tabular}{lcc}
\toprule
\textbf{Condition} & \textbf{Single} & \textbf{Multi} \\
\midrule
explicit symbol & 8 & 4 \\
explicit random word & 0 & 8 \\
implicit & 14 & 5 \\
\bottomrule
\end{tabular}
\caption{Incorrect pattern-matched square / cube answers (out of 15 possible trials)}
\label{tab:error-analysis}
\end{table}

Further, when o$1$-mini answered a problem incorrectly, its responses were often inconsistent across the five trials of that problem. Notably, in 50\% of single-character cases lacking explicit and familiar operators, all five responses were distinct and incorrect. This further shows that performance appears to depend on the presence of explicit operator cues; in their absence, o1-mini does not reliably solve the problem.












\subsection{Providing contextual information}


Our first experiment showed that in the absence of problem-specific instructions, when given a linguistic-mathematical problem directly, LLMs struggle to solve it unless the operations are both explicit and familiar. This leaves open the question of whether providing additional problem-specific information would affect the model performance. We thus modulate the context of the problem in three different ways. We query the same four problem variants as described in \Cref{tab:operation-experiment}, additionally providing the following contextual information: 

\textbf{Language:} ``Here is a puzzle \textit{based on numbers in the \{language\} language}."

\textbf{Base:} ``Here is a puzzle \textit{based on numbers in a language that uses a base-\{\textit{n}\} numeral system}."

\textbf{Implicit operations:} ``Here is a puzzle \textit{based on numbers in a language. In this language, numbers may be constructed through implicit operations like addition (twenty-nine = 20 + 9) or multiplication (five hundred = 5 $\times$ 100)}." [only for \textsc{implicit} condition]

\begin{figure}[h]
    \centering
    \includegraphics[width=\linewidth]
    {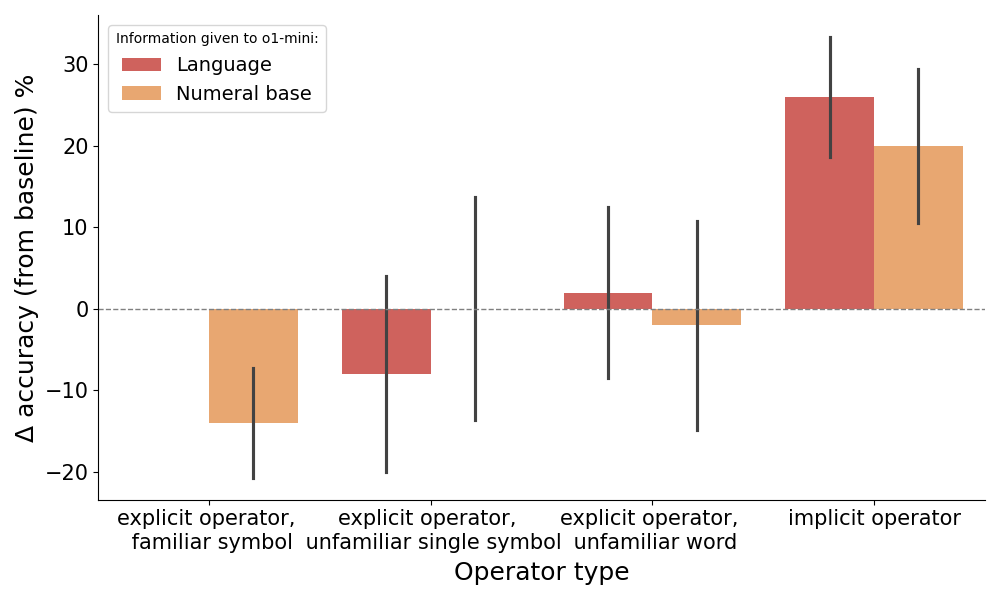}
    \caption{\textbf{Language and base information only helps in the \textsc{implicit} case.} Effect of adding language or numeral base information, plotted as a difference from the baseline values in \Cref{fig:1} for o1-mini. In cases with explicit operators, conflating overtly mathematical and linguistic information appears to confuse the models.}
    \label{fig:prompt_vars}
\end{figure}

\begin{figure}[h]
    \centering
    \includegraphics[width=\linewidth]{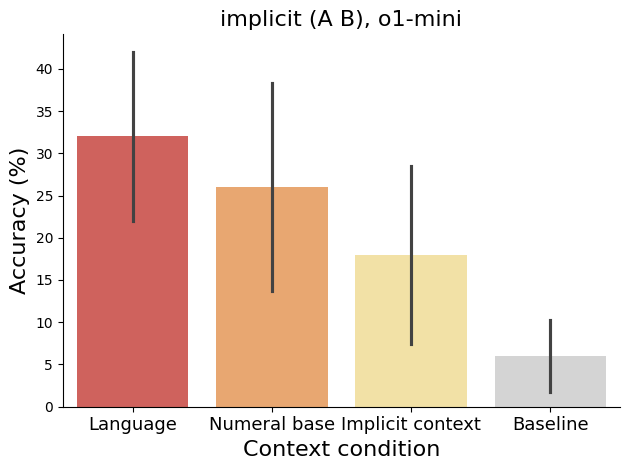}
    \caption{\textbf{Extra information improves performance on \textsc{implicit} problems (A B)}. Information about implicitness is helpful, but not as much as more direct information like the problem language. Error bars denote standard error of the mean. 5 iterations / problem.}
    \label{fig:impl_all4}
\end{figure}

We compare these to the baseline results from \Cref{expt:1} for o1-mini, presenting our results in \Cref{fig:prompt_vars}. In cases \textit{other} than the implicit operator condition, the model seems to recognize the problem as requiring a more mathematical kind of reasoning, so providing linguistic information seems to confuse the model and average performance is worse. However, in the implicit operator (A B) condition, model performance improves significantly, perhaps because the setting of the problem is less overtly mathematical. In \Cref{fig:impl_all4}, we show that providing information about the implicit reasoning needed is not as significant a boost as activating knowledge about the specific language. 

\subsection{Ablations: constructed minimal-pair problems}

In order to ensure that it is the difference in operators (as opposed to other features of the numeral system) that explains the models' inability to solve these problems, we performed an ablation study to test whether models could handle other aspects of numeral construction and combination. Our experiment is inspired by the notion of a linguistic \textit{minimal pair}, a pair of linguistic items that differ in exactly one meaningful element. We construct minimal pairs of simple, synthetic number system problems, where every element is the same except for one specific parameter that differs between two paired problems. We tested five major parameters of numeral systems, as described in \Cref{tab:minpair}.

\begin{figure}[htbp]
    \centering
    \includegraphics[width=0.7\linewidth]{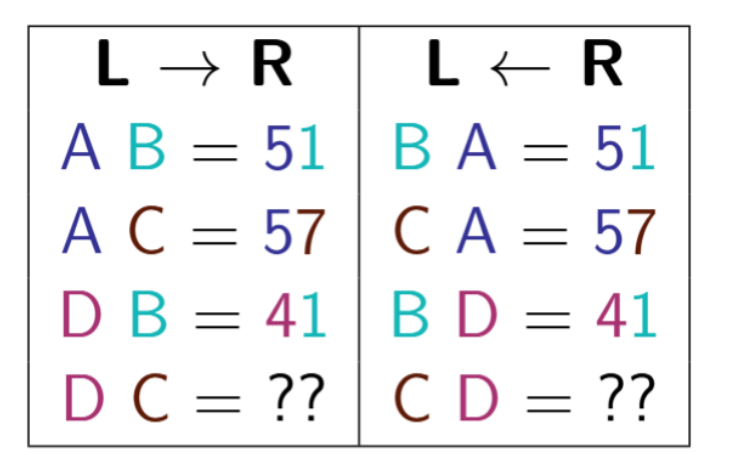}
    \caption{Example of full minimal pair template problem, for the Order parameter, where we varied whether digits are read left-to-right or right-to-left. 
    }
    \label{fig:minpair_temp}
\end{figure}

In all cases, GPT-4 and more advanced models could solve the template problems. It thus appears that most basic ``building blocks" of number systems (e.g.~the base of the system, the order of numerals, etc.) did not affect model performance in isolation, but the models consistently fail to solve number problems that involve constructing and combining complex numerals.

\begin{table}[H]
\setlength{\tabcolsep}{2pt}      
  \small
{\fontfamily{lmss}\selectfont
\scalebox{0.96}{%
  \begin{tabular}{@{}cccc@{}}
 \toprule
 \textbf{Parameter} & \textbf{GPT-$3.5$} & \textbf{GPT-$4$} & \textbf{o$1$-mini} \\ 
 \midrule 
 \makecell{Numeral system vs.\ not \\ \textcolor{BlueViolet}{A}\textcolor{TealBlue}{B} = \textcolor{BlueViolet}{fifty} \textcolor{TealBlue}{one} | \textcolor{BlueViolet}{A}\textcolor{TealBlue}{B} = \textcolor{BlueViolet}{big} \textcolor{TealBlue}{bird}}  & \textcolor{ForestGreen}{\cmark}  &  \textcolor{ForestGreen}{\cmark} &  \textcolor{ForestGreen}{\cmark}  \\ 
 \midrule
 \makecell{Typed vs.\ glyph \\  \textcolor{BlueViolet}{A}\textcolor{TealBlue}{B} = \textcolor{BlueViolet}{fifty} \textcolor{TealBlue}{one} | \textcolor{BlueViolet}{A}\textcolor{TealBlue}{B} = \textcolor{BlueViolet}{5}\textcolor{TealBlue}{1}} & \textcolor{ForestGreen}{\cmark}  &  \textcolor{ForestGreen}{\cmark} &  \textcolor{ForestGreen}{\cmark} \\
 \midrule
 \makecell{Order L $\rightarrow$ R vs. L $\leftarrow$ R\\  \textcolor{BlueViolet}{A}\textcolor{TealBlue}{B} = \textcolor{BlueViolet}{5}\textcolor{TealBlue}{1}  |    \textcolor{TealBlue}{B}\textcolor{BlueViolet}{A}  = \textcolor{BlueViolet}{5}\textcolor{TealBlue}{1}} &  \textcolor{ForestGreen}{\cmark}  &  \textcolor{ForestGreen}{\cmark} &  \textcolor{ForestGreen}{\cmark}  \\ 
 \midrule
 \makecell{Additive vs.\ subtractive \\ \makecell{\textcolor{BlueViolet}{A}\textcolor{TealBlue}{B} = \textcolor{BlueViolet}{2}\textcolor{TealBlue}{7} \\ (20 + 7)} |  \makecell{\textcolor{BlueViolet}{A}\textcolor{TealBlue}{B} = \textcolor{BlueViolet}{2}\textcolor{TealBlue}{7} \\ (30 - 3)}} & \textcolor{red}{\xmark}  &  \textcolor{ForestGreen}{\cmark} &  \textcolor{ForestGreen}{\cmark} \\ 
 \midrule
 Base of the numeral system* & \textcolor{red}{\xmark}  &  \textcolor{ForestGreen}{\cmark} &  \textcolor{ForestGreen}{\cmark} \\ 
 \bottomrule
\end{tabular}}
}
 \caption{Minimal pair results: GPT-$4$ and o1-mini solve all problem pairs, GPT-$3.5$-turbo struggles with numeral base and combination. DeepSeek-R1-distill-Qwen-7B does not produce an answer in the right format for most settings, so it is excluded from this table. Further data on testing all bases 4-19 linked in \Cref{tab:base}.}
 \label{tab:minpair}
\end{table}

\section{Discussion and Conclusions}

We study the entanglement between linguistic and numeric knowledge in language models, focusing on the ability of models to use mathematical reasoning in problems that display the implicit numerical structure in language. In the setting of these linguistic-mathematical puzzles, we show that the overtness and familiarity of operators affects the performance of language models, although many humans are able to understand how numeral systems work and hence solve the problems without needing specified operators. However, a broader study with different controls and parameter settings remains open for future work. Since all our evaluation was standardized and closed-form, we welcome research on open-ended evaluation of reasoning task responses. Current language models seem  to display some level of emergent modular structure \cite{teehan-etal-2022-emergent, lepori2023break} --- perhaps linguistic and mathematical tasks activate separate circuits or subspaces in models, and understanding the ways in which reasoning fine-tuning and reinforcement learning interacts with linguistic pretraining is another promising avenue for future research. Investigating such questions enriches our understanding of both computational and human approaches to representing numbers in language. The ability to understand language and abstract rule-governed systems is a fascinating aspect of human intelligence, and we hope that our research provides some insight into the understanding of this remarkable human trait.

\section*{Limitations}


We acknowledge the possibility that our results are explained by limitations in the training data and the small size of our dataset, as language models often equal human performance on benchmarks for which they have large quantities of similar-enough training data \cite{achiam2023gpt}. Perhaps an LLM trained on a massive corpus of linguistic number system problems would be able to solve new, previously unseen number system problems. But the data today are far too limited for such an approach, and crucially, a human solver who is familiar with existing number system problems can generalize to unseen problems extremely well! Even a human solver who is unfamiliar with existing number system problems can in theory solve any problem they are provided just by logically reasoning. Importantly, we note that although this may not be true of the \textit{average} human, when comparing the top end of humans with the top-performing current language models, it is clear that intuiting rules from human-scale data is still challenging for LLMs.


\section*{Acknowledgments}

The authors gratefully thank Tom McCoy and Kaden Holladay for helpful discussions in the initial stages of this project. This work has been made possible in part by a gift from the Chan Zuckerberg Initiative Foundation to establish the Kempner Institute for the Study of Natural and Artificial Intelligence at Harvard University.


\bibliography{custom}

\appendix

\label{sec:appendix}

\section{Randomization procedure for task-external knowledge and tokenization handling}
\label{subsec:randomize}


In this section, we address the specific changes we make across linguistic number system problems to convert them into templates suitable for our dataset. In order to truly test whether the model is \emph{solving} a problem, it should not be affected by factors external to the problem, such as flawed tokenization or the usage of memorized knowledge external to the provided task.\footnote{Memorized knowledge would also help a human solver, but people are much less likely to know the number systems of different (particularly low-resource) languages. Although linguistics olympiad contestants might know more number systems than the average person, there are over 7,000 human languages, so the probability of knowing a specific system is low. Moreover, since LLM training corpora scrape large portions of the internet, the breadth of their memorized knowledge far exceeds that of an average human.} 


Our strategy to remediate this issue is thus: in the single-letter token setting, we separated all characters by whitespaces to ensure correct tokenization. In the multi-token setting, we identified all meaningful morphemes in the problems and standardized them to remove any phonological changes, such that every morpheme had exactly one surface representation. We separated every meaningful morpheme with whitespaces, and mapped each morpheme to a randomly generated multi-token ``dummy word" for each iteration of each experiment. We created each of these ``dummy words"  by randomly sampling short tokens (length $\leq 3$) from the language models' respective tokenizer vocabularies, and concatenating tokens together to create unfamiliar words. 

For tokenizers which use schemes like byte-pair encoding, any input string will get mapped to some sequence of tokens that are present in the vocabulary, so there is no situation in which the model will see an unknown token. Since the dummy words themselves have no meaning, the model cannot directly draw on task-external linguistic information to solve the presented problems. For simplicity we restricted the random draw to those containing only romanized (Latin alphabet) characters. We also excluded tokens that contained any numeral symbols from 0-9, to ensure that the the mathematical correctness of the problems was not affected.


\section{Multi-character-variable results from Exp 1}
\label{app:multi_char}

\begin{figure}[H]
    \centering
    \includegraphics[width=1.1\linewidth]{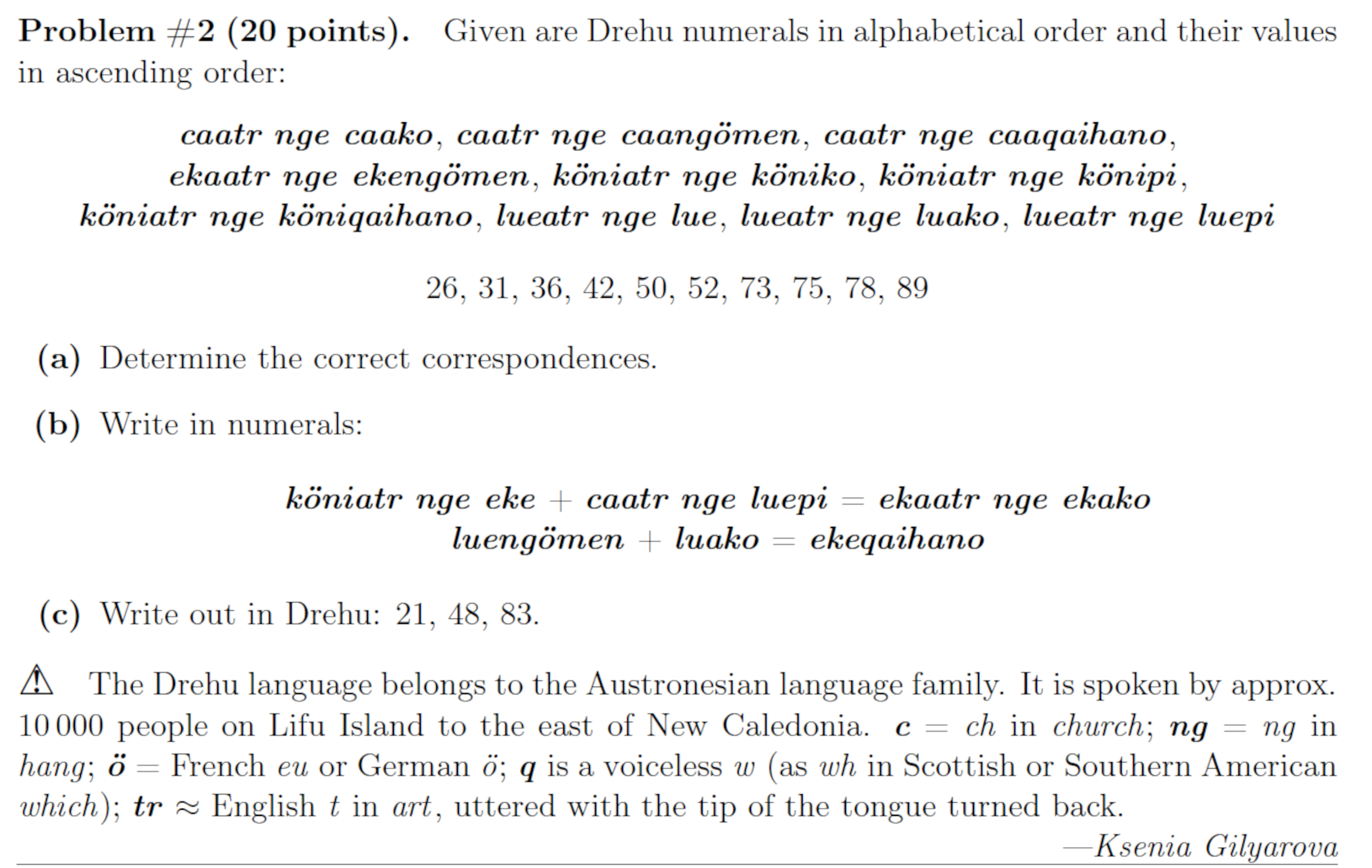}
    \caption{Drehu (IOL 2010) problem}
    \label{fig:drehu}
\end{figure}

We provide an example of our four variations of the puzzle in \Cref{tab:drehu_vars}. To query all four variants, we used the same prompt ``Here is a puzzle. Can you solve it? Please output only the answer (in place of the ??) and nothing else!". 

\begin{table}[htbp]
\small
{\fontfamily{lmss}\selectfont
\setlength\tabcolsep{0pt}
\begin{tabular}{c}
\toprule
\textbf{Explicit + familiar} \\

  \makecell{%
   (masaad $\times$ pagig) + masaad + opbob = 31 \\
   (masaad $\times$ pagig) + masaad + buylen = 26 \\
   $\vdots$ \\
   (ajssci $\times$ pagig) + (ajssci $\times$ kould) = 50 \\
   (innops $\times$ pagig) + innops + opbob = ??\\[1ex]
  } \\

  \midrule

 \textbf{Implicit} \\
    \makecell{%
   masaad pagig nge masaad opbob = 31 \\
   masaad pagig nge masaad buylen = 26 \\
   $\vdots$ \\
   ajssci pagig nge ajssci kould = 50 \\
   innops pagig nge innops opbob = ??\\[1ex]
  }  \\

\midrule

\textbf{Explicit + unfamiliar (Greek)}
 \\[1ex]

 \makecell{%
   (masaad $\beta$ pagig) $\alpha$ masaad $\alpha$ opbob = 31 \\
   (masaad $\beta$ pagig) $\alpha$ masaad $\alpha$ buylen = 26 \\
   $\vdots$ \\
   (ajssci $\beta$ pagig) $\alpha$ (ajssci $\beta$ kould) = 50 \\
   (innops $\beta$ pagig) $\alpha$ innops $\alpha$ opbob = ??
  } \\ 

  \midrule

  \textbf{Explicit + unfamiliar (random)} \\

      \makecell{%
   (masaad hibcat pagig) xebrut masaad xebrut opbob = 31 \\
   (masaad hibcat pagig) xebrut masaad xebrut buylen = 26 \\
   $\vdots$ \\
   (ajssci hibcat pagig) xebrut (ajssci hibcat kould) = 50 \\
   (innops hibcat pagig) xebrut innops xebrut opbob = ??
  } \\

\bottomrule
\end{tabular}
}
\caption{Example of four problem variants in the multi-character setting, corresponding to Drehu (IOL 2010) dataset problem in \Cref{fig:drehu}.}
\label{tab:drehu_vars}
\end{table}

\section{Base experiment}

\begin{figure}[h]
  \includegraphics[width=\linewidth]{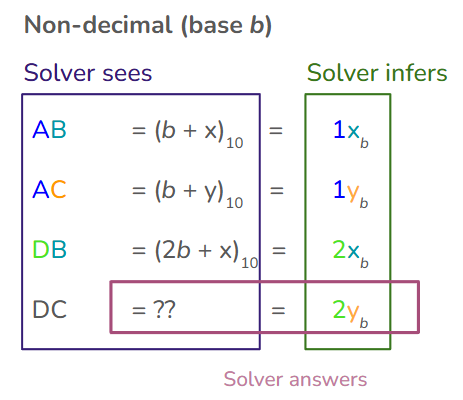}
  \caption {Setup for base experiment}
  \label{fig:base_expt}
\end{figure}

In order to understand whether sufficiently advanced language models would show performance that was invariant to changes in the base, we conducted a more fine-grained minimal pair experiment into the effect of numeral base on problem performance. Here, the solver would see the Hindu-Arabic numerals corresponding to the English base-10 representation of the numbers, because the problem was presented in English. But the unknown symbols corresponded to the numbers as expressed in a different base, as shown in \Cref{fig:base_expt}.

We conducted two different versions of this experiment. First, we mapped the unknown symbols to the single-character whitespaced \textit{A, B, C}, and \textit{D} tokens, as in \Cref{fig:base_expt}. In the second version, each of the four unknown symbols (\textit{A, B, C, D}) was instead represented by a corresponding random token drawn from the tokenizer vocabulary, to ensure that the context of the specific tokens \textit{A, B, C}, and \textit{D} was not influencing our results. 

We tested four increasingly sophisticated GPT models (GPT-$3.5$-turbo, GPT-$4$, GPT-$4$o, and o$1$-mini) on both versions of the experiment and provide results in \Cref{tab:base}. GPT-4o and o1-mini solved all problems in both conditions, displaying performance that was robust to the base of the problem. 


\section{Table of languages}
\label{app:lang-table}

The 10 problems that we used for our analyses. The problems range in difficulty from the first and second rounds of the UK Linguistics Olympiad (UKLO R1 and R2) to the International Linguistics Olympiad, which typically has the most challenging problems. 
\begin{table}[H]
\setlength{\tabcolsep}{2.5pt}      
  \begin{tabular}{@{}ccccc@{}}
 \toprule
 \textbf{Language} &  \textbf{ISO code} & \textbf{Base}  & \textbf{Level} \\ 

 \midrule 
 Drehu & dhv & 20   & IOL  \\
 Georgian & kat & 20  & UKLO R1\\ 
 Gumatj & gnn & 5 & UKLO R1 \\ 
 Ndom & nqm & 6  & IOL \\ 
 Ngkolmpu & kcd & 6  & UKLO R1 \\ 
 Northern Pame & pmq & 8 & UKLO R1\\ 
 Umbu-Ungu & ubu & 24  & IOL \\ 
 Waorani & auc & 5   & UKLO R1 \\ 
 Yoruba & yor & 20  & UKLO R2 \\ 
 Yup'ik & esu & 20   & UKLO R2  \\ 
      
 \hline
\end{tabular}
 \caption{Languages and problem features in final dataset (after removing/standardizing phenomena)}
 \label{tab:probs}
\end{table}

\begin{table*}[h]
\centering
\begin{tabular}{ccccccccc}
\toprule
\textbf{Base} & \multicolumn{2}{|c}{\textbf{GPT-$3.5$-turbo}} & \multicolumn{2}{c}{\textbf{GPT-$4$}} & \multicolumn{2}{c}{\textbf{GPT-$4$o}} & \multicolumn{2}{c}{\textbf{o$1$-mini}} \\
\midrule
{}   & \textbf{ABCD} & \textbf{Random} & \textbf{ABCD} & \textbf{Random} & \textbf{ABCD} & \textbf{Random} & \textbf{ABCD} & \textbf{Random} \\
\midrule
4    &  \textcolor{red}{\xmark} &  \textcolor{ForestGreen}{\cmark}   & \textcolor{ForestGreen}{\cmark} &  \textcolor{ForestGreen}{\cmark}  & \textcolor{ForestGreen}{\cmark} &  \textcolor{ForestGreen}{\cmark} & \textcolor{ForestGreen}{\cmark} &  \textcolor{ForestGreen}{\cmark} \\
5    &  \textcolor{red}{\xmark} &  \textcolor{red}{\xmark}   & \textcolor{ForestGreen}{\cmark} &  \textcolor{ForestGreen}{\cmark}  & \textcolor{ForestGreen}{\cmark} &  \textcolor{ForestGreen}{\cmark} & \textcolor{ForestGreen}{\cmark} &  \textcolor{ForestGreen}{\cmark} \\
6    &  \textcolor{ForestGreen}{\cmark} &  \textcolor{red}{\xmark}  & \textcolor{ForestGreen}{\cmark} &  \textcolor{ForestGreen}{\cmark}  & \textcolor{ForestGreen}{\cmark} &  \textcolor{ForestGreen}{\cmark} & \textcolor{ForestGreen}{\cmark} &  \textcolor{ForestGreen}{\cmark} \\
7    &  \textcolor{ForestGreen}{\cmark} &  \textcolor{ForestGreen}{\cmark}   & \textcolor{ForestGreen}{\cmark} &  \textcolor{ForestGreen}{\cmark}  & \textcolor{ForestGreen}{\cmark} &  \textcolor{ForestGreen}{\cmark} & \textcolor{ForestGreen}{\cmark} &  \textcolor{ForestGreen}{\cmark} \\
8    &  \textcolor{red}{\xmark} &  \textcolor{red}{\xmark}   & \textcolor{ForestGreen}{\cmark} &  \textcolor{ForestGreen}{\cmark}  & \textcolor{ForestGreen}{\cmark} &  \textcolor{ForestGreen}{\cmark} & \textcolor{ForestGreen}{\cmark} &  \textcolor{ForestGreen}{\cmark} \\
9    &  \textcolor{ForestGreen}{\cmark} &  \textcolor{ForestGreen}{\cmark}   & \textcolor{ForestGreen}{\cmark} &  \textcolor{ForestGreen}{\cmark}  & \textcolor{ForestGreen}{\cmark} &  \textcolor{ForestGreen}{\cmark} & \textcolor{ForestGreen}{\cmark} &  \textcolor{ForestGreen}{\cmark} \\
11    &  \textcolor{ForestGreen}{\cmark} &  \textcolor{ForestGreen}{\cmark}   & \textcolor{ForestGreen}{\cmark} &  \textcolor{ForestGreen}{\cmark}  & \textcolor{ForestGreen}{\cmark} &  \textcolor{ForestGreen}{\cmark} & \textcolor{ForestGreen}{\cmark} &  \textcolor{ForestGreen}{\cmark} \\
12   &  \textcolor{red}{\xmark} &  \textcolor{ForestGreen}{\cmark}   & \textcolor{red}{\xmark} &  \textcolor{ForestGreen}{\cmark}  & \textcolor{ForestGreen}{\cmark} &  \textcolor{ForestGreen}{\cmark} & \textcolor{ForestGreen}{\cmark} &  \textcolor{ForestGreen}{\cmark} \\
13   &  \textcolor{red}{\xmark} &  \textcolor{red}{\xmark}   & \textcolor{ForestGreen}{\cmark} &  \textcolor{ForestGreen}{\cmark}  & \textcolor{ForestGreen}{\cmark} &  \textcolor{ForestGreen}{\cmark} & \textcolor{ForestGreen}{\cmark} &  \textcolor{ForestGreen}{\cmark} \\
14   &  \textcolor{red}{\xmark} &  \textcolor{red}{\xmark}   & \textcolor{ForestGreen}{\cmark} &  \textcolor{ForestGreen}{\cmark}  & \textcolor{ForestGreen}{\cmark} &  \textcolor{ForestGreen}{\cmark} & \textcolor{ForestGreen}{\cmark} &  \textcolor{ForestGreen}{\cmark} \\
15   &  \textcolor{red}{\xmark} &  \textcolor{ForestGreen}{\cmark}   & \textcolor{ForestGreen}{\cmark} &  \textcolor{ForestGreen}{\cmark}  & \textcolor{ForestGreen}{\cmark} &  \textcolor{ForestGreen}{\cmark} & \textcolor{ForestGreen}{\cmark} &  \textcolor{ForestGreen}{\cmark}  \\
16   &  \textcolor{ForestGreen}{\cmark} &  \textcolor{red}{\xmark}   & \textcolor{red}{\xmark} &  \textcolor{ForestGreen}{\cmark}  & \textcolor{ForestGreen}{\cmark} &  \textcolor{ForestGreen}{\cmark} & \textcolor{ForestGreen}{\cmark} &  \textcolor{ForestGreen}{\cmark}  \\
17   &  \textcolor{red}{\xmark} & \textcolor{red}{\xmark}  & \textcolor{ForestGreen}{\cmark} &  \textcolor{ForestGreen}{\cmark}  & \textcolor{ForestGreen}{\cmark} &  \textcolor{ForestGreen}{\cmark} & \textcolor{ForestGreen}{\cmark} &  \textcolor{ForestGreen}{\cmark}  \\
18   &  \textcolor{ForestGreen}{\cmark} &  \textcolor{ForestGreen}{\cmark}   & \textcolor{ForestGreen}{\cmark} &  \textcolor{ForestGreen}{\cmark}  & \textcolor{ForestGreen}{\cmark} &  \textcolor{ForestGreen}{\cmark} & \textcolor{ForestGreen}{\cmark} &  \textcolor{ForestGreen}{\cmark}  \\
19   &  \textcolor{ForestGreen}{\cmark} &  \textcolor{red}{\xmark}   & \textcolor{ForestGreen}{\cmark} &  \textcolor{ForestGreen}{\cmark}  & \textcolor{ForestGreen}{\cmark} &  \textcolor{ForestGreen}{\cmark} & \textcolor{ForestGreen}{\cmark} &  \textcolor{ForestGreen}{\cmark} \\
\bottomrule
\end{tabular}
 \caption{Base experiment results: GPT-$4$o / o$1$-mini solve every problem, regardless of randomization}
 \label{tab:base}
\end{table*}

\begin{figure*}[bp]

  \includegraphics[width=\linewidth]{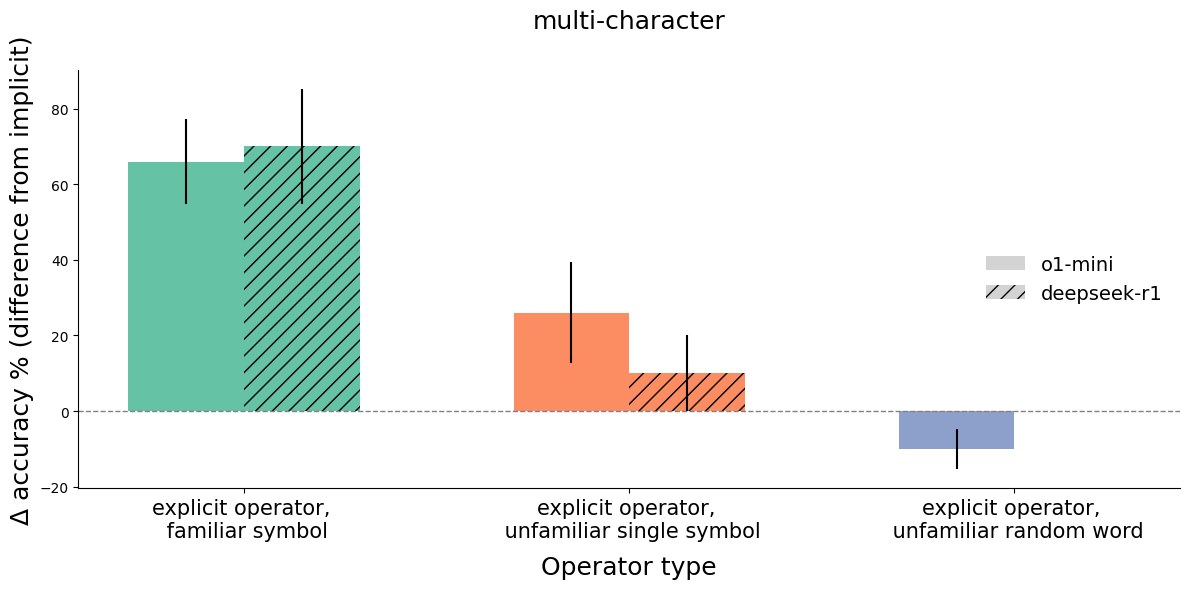}
  
  \caption {Both o1-mini and DeepSeek struggle with the explicit-unfamiliar condition (o1 shows negative improvement, DeepSeek shows 0\%) in the multi-character setting. Error bars denote standard error of the mean. 5 iterations / problem tested for 10 problems.}

  \label{fig:multitok_delta}
\end{figure*}

\begin{figure*}[t]

\includegraphics[width=\linewidth]{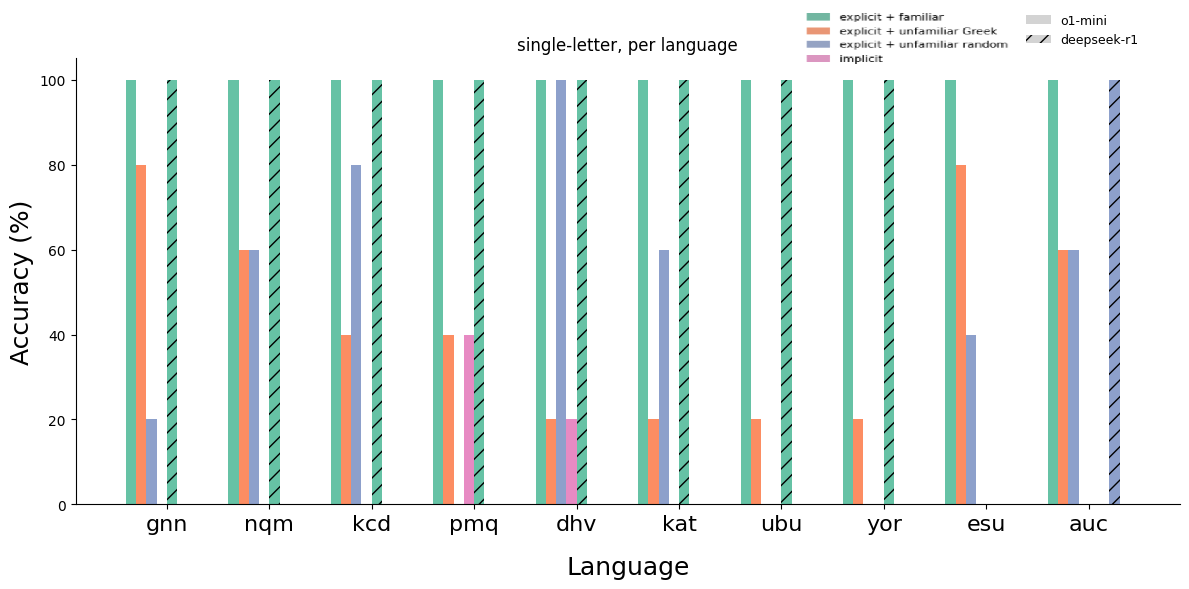}
  \hfill  \includegraphics[width=\linewidth]{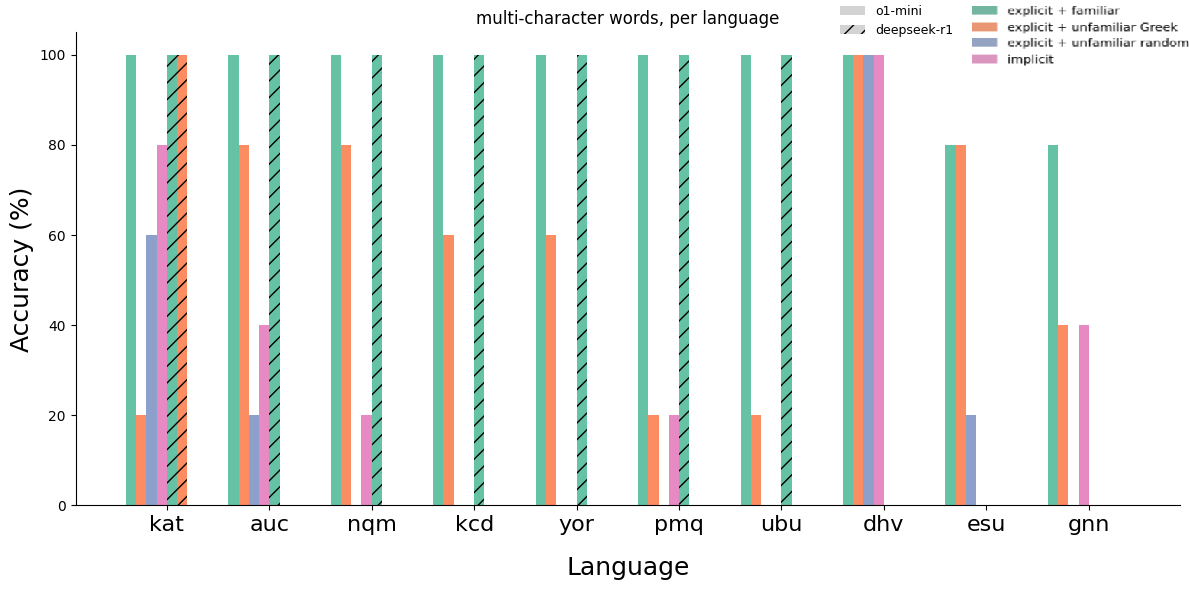}
  \caption {\textbf{Results per language, (a) single-character (b) multi-character: performance varies significantly by problem and operator type.} Note that Drehu (dhv) and Georgian (kat) are two of the easiest problems in our dataset: much of the difficulty for human solvers is in the phonological changes and unintuitive order that the numbers are presented in, both of which we standardize away for our controlled datasets. Without those parameters, the systems are straightforward vigesimal-decimal systems like French, which the models have almost certainly had exposure to.}
\end{figure*}


\end{document}